# Antisocial behavior towards large language model users: experimental evidence

Paweł Niszczota [1,3,*] and Cassandra Grützner [2,1,3]

[1] Poznań University of Economics and Business, Humans & Artificial Intelligence Laboratory (HAI Lab), Institute of International Business and Economics, al. Niepodległości 10, 61-875 Poznań, Poland; pawel.niszczota@ue.poznan.pl

[2] Martin Luther Universität Halle-Wittenberg, Chair for Business Ethics & Management Accounting, Große Steinstraße 73, 06108 Halle, Germany, cassandra.gruetzner@wiwi.uni-halle.de

[3] Equal contributions.

* Corresponding author: Paweł Niszczota, Poznań University of Economics and Business, al. Niepodległości 10, 61-875, Poznań, Poland; telephone number: +48 61 854 33 24; pawel.niszczota@ue.poznan.pl; ORCID: 0000-0002-4150-3646

## Statements and Declarations

**Competing interests.** The authors have no relevant financial or non-financial interests to disclose.

**Ethics approval.** Approval was obtained from the Committee of Ethical Science Research at the Poznań University of Economics and Business (resolution 1/2023). The procedures used in this study adhere to the tenets of the Declaration of Helsinki.

**Consent.** Informed consent was obtained from all individual participants included in the study.

**Data, materials, and code availability.** The pre-registration documents, data, materials, and code are available at: https://osf.io/2yt9e/overview?view_only=8268c7ac95d94f2390fd3ed6dc76269a

## Acknowledgments

This research was supported by grant 2021/42/E/HS4/00289 from the National Science Centre, Poland.

**Antisocial behavior towards large language model users: experimental evidence**

**Abstract**

The rapid spread of large language models (LLMs) has raised concerns about the social reactions they provoke. Prior research documents negative attitudes toward AI users, but it remains unclear whether such disapproval translates into costly action. We address this question in a two-phase online experiment ($N = 491$ Phase II participants; Phase I provided targets) where participants could spend part of their own endowment to reduce the earnings of peers who had previously completed a real-effort task with or without LLM support. On average, participants destroyed 36% of the earnings of those who relied exclusively on the model, with punishment increasing monotonically with actual LLM use. Disclosure about LLM use created a credibility gap: self-reported null use was punished more harshly than actual null use, suggesting that declarations of "no use" are treated with suspicion. Conversely, at high levels of use, actual reliance on the model was punished more strongly than self-reported reliance. Taken together, these findings provide the first behavioral evidence that the efficiency gains of LLMs come at the cost of social sanctions.

# 1. Introduction

The rapid integration of generative AI into workplaces promises a new era of automation, ostensibly freeing human cognition for more complex and creative tasks. This narrative, however, overlooks a fundamental tension: as AI systems become more capable of automating human effort, human agency is increasingly challenged—not by the machines themselves, but by the social reactions of other humans. With insights from the industry showcasing that AI-users are increasingly concerned about being perceived lazy (Asana, 2024), and a recent study by Reif et al. (2025) indicating that these worries are not unfounded, a new form of social policing is emerging: the decision to offload cognitive labor to an AI can trigger negative judgments from peers. This tension places human agency in a double bind: individuals are encouraged to use powerful tools for efficiency, yet risk social sanction for doing so, effectively constraining their agency by the social networks they are part of.

Existing research has documented such negative evaluations: AI users are perceived as less competent and less motivated (Reif et al., 2025), and even disclosing AI assistance can lead to harsher assessments (Cheong et al., 2025). This suggests a troubling misalignment between the ethical imperative of transparency and the lived social reality that transparency may invite punishment. Yet these studies capture primarily attitudes. A critical question remains unanswered: are negative judgments strong enough to translate into costly actions? Do people merely disapprove of AI users, or are they willing to sacrifice their own resources to punish them, thereby enforcing a social norm that privileges unaided human effort?

This study addresses this gap by moving from attitudes to behavior. Using an experimental economic paradigm—a variant of the Money Burning Game (Zizzo, 2003)—we test whether individuals will incur a personal cost to reduce the monetary gains of peers who used a large language model (LLM) to complete a real-effort task. By manipulating whether AI use was observable or self-reported, we probe when transparency builds trust versus when it backfires. In doing so, we provide the first behavioral evidence of costly antisocial responses to AI use, directly operationalizing the tension between the drive for efficiency through automation and the protection of human agency.

To explain why the use of AI tools might provoke costly punishment, we draw upon two complementary theoretical frameworks: one centered on fairness and effort-based deservingness, and another on signaling and the credibility of disclosure.

## 2. Theoretical underpinnings and hypotheses

### 2.1. Inequity Aversion and Costly Punishment

The propensity to punish others for perceived unfairness, even at a personal cost, is a well-documented phenomenon in behavioral economics. Our experimental approach is grounded in this literature, which provides our basis for expecting antisocial behavior towards individuals who leverage AI to gain an advantage.

A fundamental principle in social interactions is the norm of effort-based deservingness: the belief that rewards should be proportional to personal effort (Feather, 1999). Gains achieved through shortcuts or unearned advantages are often perceived as illegitimate, violating notions of a just world (Lerner, 1980). The use of a capable LLM to complete a real-effort task represents a quintessential violation of this norm. It decouples reward from individual exertion, allowing a user to attain outcomes with significantly less personal effort than a peer who works unaided.

A growing body of recent empirical evidence confirms that this violation triggers significant social and reputational penalties. Lim & Schmälzle (2024) show that people explicitly view outcomes achieved with AI as less deserved and the advantage as less legitimate, directly linking these judgments to just-world and deservingness frameworks. Similarly, Earp et al. (2024) and He et al. (2025) document that AI users receive less credit and reputational recognition, even when performance improves—evidence that social attributions penalize visible shortcuts. Consequently, the act of using an LLM is not merely a technical shift but a social one, directly challenging the effort-reward link that underpins judgments of fairness, legitimacy, and deservedness.

This perception of illegitimacy can trigger a motivation to punish, rooted in preferences for fairness and inequity aversion. As formalized by Fehr and Schmidt (1999), individuals experience disutility from disadvantageous inequality—situations where others are perceived to have gained more from an unfair process. This aversion can be strong enough to motivate costly actions to reduce others' advantages, even in the absence of direct competition. The Money Burning Game Zizzo (2003) precisely captures this propensity, showing that people are willing to sacrifice their own resources to destroy wealth they deem unfairly acquired.

Although money burning is a mechanism to counter wealth accumulation that might seem unfair, similar types of games – Joy of Destruction Games (Abbink & Herrmann, 2011; Abbink & Sadrieh, 2009) – suggest that people might be willing to perform money burning even when there are no motivations for doing so. While some people are willing to reduce others' wealth simply indicates antisocial behavior, we assume that the destruction of others' wealth is mostly a means of combating unfairness consistent with antisocial behavior. Therefore, to appropriately assess the effect of LLM-use on burning behavior, we will compare each use-case against the case when a person did not have access to an LLM at all. We posit that:

**H1: Use of LLMs by people increases antisocial behavior towards them relative to people who did not have access to LLMs.**

Importantly, prior work suggests that punishment is not an all-or-nothing response but scales with the perceived severity of a violation. Norm enforcement theory holds that stronger deviations from expected behavior elicit stronger sanctions (Bicchieri, 2006) Fairness theories similarly emphasize proportionality between effort and reward (Konow, 2003), implying that as AI use increases, the decoupling between effort and outcome becomes more glaring. Experimental economics confirms this logic: sanctions tend to escalate with the size of the undeserved advantage (Falk et al., 2005). AI-specific research corroborates these dynamics. Lim & Schmälzle (2024) show that heavy reliance on AI is judged as less legitimate than occasional use; Acar et al. (2025) find reputational penalties grow with intensity of use; and Reif et al. (2025) report harsher social evaluations for extensive users compared to minimal users. Taken together, these findings suggest that punishment will increase monotonically with actual LLM use, as heavier reliance constitutes a more severe breach of the effort–reward principle.

**H2: Antisocial behavior towards LLM-users increases monotonically with the intensity of actual use.**

## 2.2. From fairness to signaling: the transparency dilemma

Crucially, fairness-based punishment presumes that observers can *see* the shortcut being taken. Yet in many real-world settings, AI use is not directly observable, but instead disclosed by the user – or left unmentioned. This reallocation of credit and blame is tied to the visibility of the shortcut: when AI use is disclosed, it can actively erode trust in both the output and the individual, creating social situations in which honesty about human-AI collaboration invites stigma (Draxler et al., 2024; Schilke & Reimann, 2025) perceptions are not confined to workplaces; in education, AI use is culturally framed as "cheating" and "lazy" (Inside Higher Ed, 2024; The Guardian, 2025; Michigan Virtual, 2024), reinforcing the idea that it confers an unearned advantage.

Signaling theory (Connelly et al., 2011; Spence, 1973) provides a useful lens for understanding these dynamics. In an environment of information asymmetry, where a worker's true level of effort (or AI use) is unobservable, any declaration – especially a declaration of *non-use* – functions as a costly signal. A declaration of null use is a claim of pure, individual effort. But signals are only effective when they are credible. If observers believe that AI use is widespread and potentially advantageous, they may view a self-report of "no AI use" with skepticism, suspecting it to be a strategically manipulated, dishonest signal (Jungbauer & Waldman, 2023) from someone who actually used the tool and wishes to avoid the associated stigma (Gonçalves et al., 2025).

In our context, self-reporting *null use* of AI, when its use is possible, may be interpreted not as virtuous restraint but as a suspicious denial – suggesting underreporting. Observed non-use, by contrast, is credible and may even be rewarded as a sign of effort. This credibility gap explains why self-reported null use should be punished more harshly than actual null use.

**H3: Self-reported null use will cause more antisocial behavior than actual null use.**

We predict that individuals who signal null use will be punished *more* than those who are genuine null users but do not make a declaration. This is because the signal itself raises the stakes, making the observer actively assess its truthfulness. A declaration of null use in a climate of distrust is a risky signal that can backfire, attracting punishment from observers who disbelieve it.

Furthermore, we expect that the relationship between *actual* use and punishment will differ from the relationship between *reported* use and punishment. Punishment based on actual use reflects a direct response to the norm violation. In contrast, punishment based on a report is a response to a *signal*, which carries its own credibility weight. The slope of antisocial behavior may be steeper for reported use if high levels of reported use are seen as brazen honesty about a major violation, or it may be shallower if honesty mitigates some of the stigma. This hypothesis tests the unique punitive weight of the signal itself, independent of the underlying behavior.

**H4. Actual and self-reported use increase antisocial behavior at different rates.**

## 3. Methodology

### 3.1. Experimental design

The experiment consisted of two phases. In Phase I, participants completed a series of real-effort tasks under one of three conditions: (1) no access to an LLM (control), (2) access to an LLM with actual usage visible to others, or (3) access to an LLM with usage privately reported by the participant. In Phase II, a new group of participants played a variant of the Money Burning Game (Zizzo, 2003; Zizzo & Oswald, 2001) against Phase I participants from all three conditions.

#### 3.1.1. Phase I

Phase I participants were asked to count the number of occurrences of a target item in a grid (e.g., frogs among distractor images such as broccoli). The three experimental conditions were as follows:

1. **Control condition** ($N = 4$): participants solved all tasks without LLM support.
2. **Actual LLM use condition** ($N = 89$): participants could consult an LLM's solution on each task, and knew that the number of tasks (0–5) for which the LLM button was used will be observable to Phase II players.
3. **Self-reported LLM use condition** ($N = 50$): participants could consult the LLM, and had to self-report their frequency of use, knowing that Phase II players were informed only of this self-report.

Note that the number of participants in each condition was not the same. As our main focus lies on the actual behavior of Phase II participants, we only recruited Phase I participants until we had at least one in each possible use case, to both remain economically prudent and non-deceptive, as participants in Phase II were truthfully informed that they would judge a person in a particular use case. However, the unequal number of participants in each condition does not affect the validity of the findings.

### 3.1.2. Phase II

Phase II employed a within-subjects design. Each participant first reviewed the five real-effort tasks used in Phase I to familiarize themselves with the effort required. Afterwards, they made Money Burning decisions against targets from all three Phase I conditions. For LLM users, deciders observed all possible intensities of use (0–5 tasks solved with LLM assistance), either actual or self-reported. Together with the control condition, this yielded 13 distinct target cases. The order of experimental blocks, i.e. the actual LLM use or the self-reported LLM use condition, was randomized, and Phase I targets were re-used across deciders to populate all pairings.

### 3.2. Real-effort task

All Phase I participants had to complete the real-effort tasks, while Phase II participants were only shown the same tasks. This ensured that deciders could gauge the task's difficulty, potentially shaping their judgments of Phase I players' reliance on the LLM.

The real-effort task was designed as follows: participants were instructed to count the occurrences of a specific emoji within a grid of distractor emojis. In the two LLM conditions, participants were informed that they could access support from a reliable and pre-tested LLM system by clicking a button. To ensure comprehension, they were presented with several simple understanding questions about the task rules. Any incorrect answers triggered a re-explanation, and participants could only proceed once all questions were answered correctly. They then completed a trial task, in which they had to count emojis in a small 3 × 3 grid, before moving on to the incentivized tasks. Finally, participants were presented with the five emoji-counting tasks which increased in difficulty, each of which offered the opportunity to earn a performance-based bonus (see **Fig. 1**).

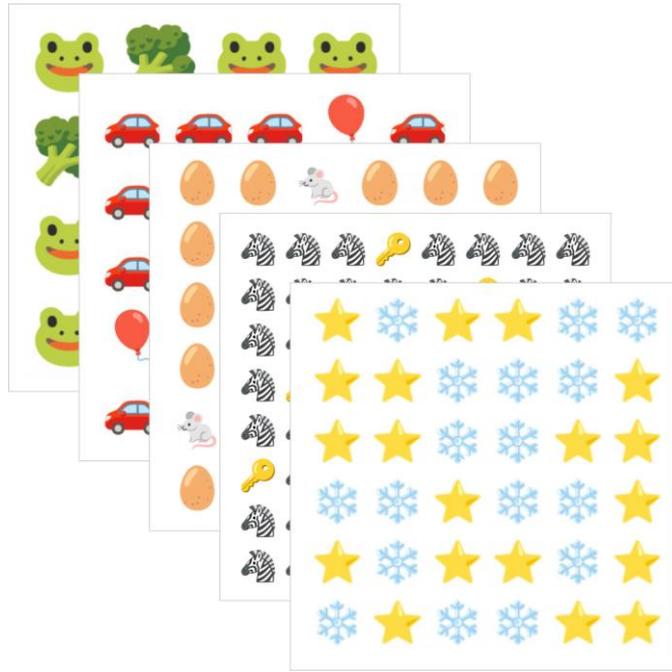

**Fig. 1. Five real-effort tasks used in the experiment**

*Notes:* In the top-left corner we present the first counting task shown to participants.

LLM support was generated in advance by prompting a frontier LLM (Google Gemini; Gemini Team Google, 2025) to count the specified emoji within each grid. The output was then presented to participants as a static image rather than generated in real time. At the time of the study, many LLM systems still struggled with image processing and emoji counting, and even Gemini did not consistently provide correct solutions. To avoid introducing errors into the experimental manipulation, we restricted the set of tasks to those that Gemini had solved correctly, thereby ensuring that participants always received accurate LLM advice.

### 3.2.1. Incentives in Phase I

Phase I participants received a flat fee of £1.00 plus £0.20 for each correctly solved task (up to £1.00). Final payoffs were subject to Phase II money burning decisions. Thus, total earnings ranged from £1.00 (flat fee only) to £2.00 (flat fee plus full bonus), but never below the flat fee.

### 3.2.2. Incentives in Phase II

Phase II participants received a flat fee of £2.25 plus an endowment equivalent to £0.40 (40 tokens, given that 1 token = £0.01). This endowment could be used to reduce the Phase I targets' bonus payments. Importantly, participants were informed that no monetary benefit accrued to deciders (or other participants) from burning, emphasizing the purely antisocial nature of the choice. Decisions were elicited using the strategy method: participants made choices for all 13 cases, with one implemented at random. Participants were told that the part of the endowment that they retained will be added to their final compensation.

### 3.3. Money Burning

Phase II deciders were informed that Phase I targets had previously completed the real-effort task. To standardize outcomes, only targets with perfect scores (eligible for the full £1.00 bonus = 100 tokens) were included. This ensured that any reduction in payoff could be attributed to perceptions of (un)deservedness rather than task performance.

For each target, deciders could spend their 40 token endowment to destroy part or all of the target's 100 token bonus. Every token spent reduced the target's payoff by 5 tokens, allowing full elimination of the 100 token bonus for 20 tokens (half of the Phase II endowment). The order of experimental conditions presented was randomized, and within the LLM-use blocks, the order of the intensity of use (0–5 LLM use) shown was also randomized. Given fewer Phase I than Phase II participants, each target could be matched with multiple deciders; in such cases, one outcome was randomly implemented for the target (Zizzo, 2003).

The task given to participants constituted a variant of the Money Burning Game (Zizzo, 2003, 2025; Zizzo & Oswald, 2001), where participants have to decide whether they are willing to use their own money to influence the payment of another person. Motives for "burning" the money of others vary. We designed the experiment so that the main motivation to burn money is to reduce the payment of another person that accumulated the money in an undeserving manner, by looking up the estimate provided by a capable LLM. In **Fig. 2** we visualize key components of our experimental design.

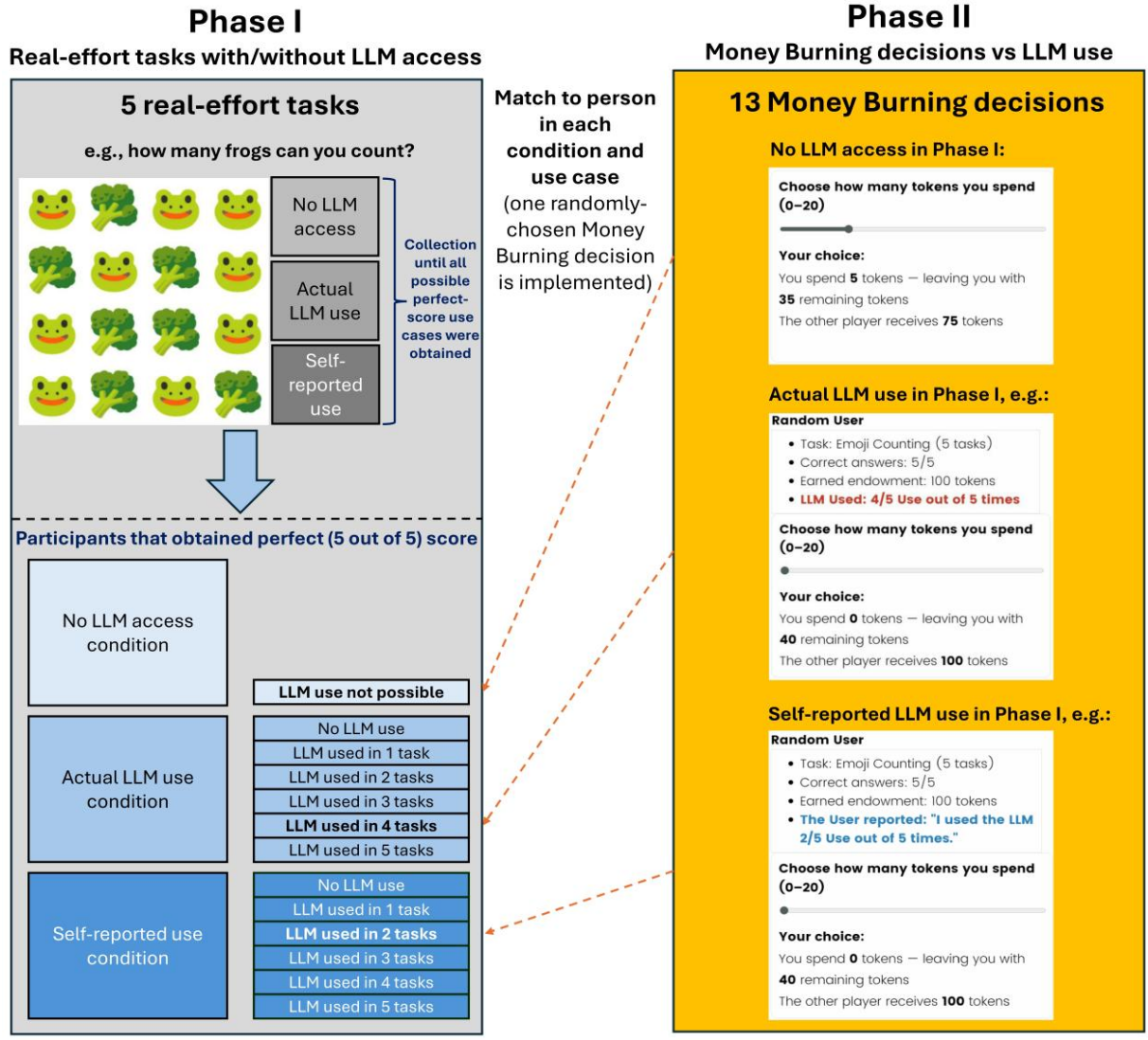

**Fig. 2. Experimental design**

*Notes:* Conditions in Phase I have an unequal number of participants, as we collected data until we had at least one perfect-scorer in each possible use case within this condition.

See Appendix for experimental instructions. We also included printouts of the experimental instructions for Phases I and II on OSF (see *Open science practices* section).

3.4. Participants

We recruited 143 participants in Phase I of the experiment, and 501 participants in Phase II. In Phase I, we recruited participants from Prolific, that had a 98% or higher approval rating, were located and born in the UK, and whose first language was English. For Phase II, we recruited a representative sample of people from the UK, again using Prolific. We excluded 10 participants who either failed both attention checks or for whom demographic data was unavailable. This left us with a final sample of 491 participants ($M_{age}$ = 46.8 (*SD* = 15.3); 51.3% female, 48.7% male). However, in the Appendix we report results for confirmatory analyses obtained if we exclude participants that did not correctly answer either of the two attention checks: however, these results are qualitatively the same. In each block, we also asked two comprehension questions, with four questions overall. The mean score of participants was 75.2%; 92.6% of participants correctly answered at least one of the comprehension questions within each block. Overall, participants thus had a good understanding of the tasks.

### 3.5. Statistical analysis

#### 3.5.1. Variables

The dependent variable was the proportion of the bonus of the participant from Phase I that was burned by the participant from Phase II. Given that each participant in Phase II was matched with a random person across the entire spectrum of use cases, there were 13 decisions per participant (1 in the control condition, 6 in the actual LLM-use condition (given the 0-5 actual use intensity range), and 6 in the self-reported LLM-use condition (given the 0-5 self-reported intensity range).

We also used several variables as control variables and potential moderators in exploratory analyses. Apart from collecting typical demographic data (gender, age) we gathered a number of variables relating to their technological affinity (nine items; Franke et al., 2019), regularity of LLM use (one item: "I frequently use LLM systems"), and knowledge about LLMs (two items: "I have a good understanding of the capacity of LLM systems." and "I have a good understanding of limitations of LLM systems."). LLM use and knowledge were very highly correlated ($r$ = 0.77, 95% CI [0.73, 0.80], $t(489)$ = 26.62, $p$ < .001), so we opted to only use the former in key regression analyses. The correlation between technological affinity and LLM use was much less substantial ($r$ = 0.39, 95% *CI* [0.31, 0.46], $t(489)$ = 9.25, $p$ < .001). The correlation between technological affinity and LLM knowledge was, similarly, less substantial ($r$ = 0.44, 95% *CI* [0.37, 0.51], $t(489)$ = 10.82, $p$ < .001).

### 3.5.2. Modelling techniques

Given that the dependent variable was a proportion, we opted to use beta regressions (Cribari-Neto & Zeileis, 2010). Our experiment had a within-subject design, and thus we used fitted beta regression mixed-models using the *glmmTMB* package in *R* (McGillycuddy et al., 2025). The dependent variable was transformed in accordance with the recommendations of Smithson and Verkuilen (2006), so that it was in an open range (did not include 0 and 1), thus becoming appropriate for the beta regression. Model 1 was used to test Hypothesis 1 and Model 2 was used to test Hypotheses 2-4 (see pre-registration file for exact specifications). Models included random intercepts for participants, random intercepts for block-order. Additionally, in Model 2 we also included a random slope for intensity.

### 3.6. Open science practices

The experiment was pre-registered at: https://aspredicted.org/9w89-d2nh.pdf. We made one deviation from the pre-registration. As noted earlier, while we pre-registered that participants had to correctly answer both attention checks, ultimately, we decided to use more lax restriction criterion, and exclude those who failed both attention checks. One of the attention checks was not correctly answered by 44.9% of participants, which is more indicative of the check being too difficult (or perhaps open to interpretations). Crucially, the results are qualitatively the same when using the restriction criterion. The revised exclusion criterion takes fuller advantage of the representative nature of the sample.

Data, code, and materials are available on the Open Science Framework at: https://osf.io/5b4g9/?view_only=ad5ae2fc0557467da72c6d7447646371.

## 4. Results

### 4.1. Descriptive statistics

We present descriptive statistics for key variables in
**Table** 1. Consistent with prior work showing that people might be willing to engage in wealth destruction even in the absence of a rationale for doing so (Abbink & Herrmann, 2011), there was money burning even for people that did not use an LLM in the real-effort tasks. More specifically, the mean money burning was 9.65% of the maximum amount that could be burned by a participant in Phase II,

destroying 9.65% of the earnings of a participant from Phase I). This served as the baseline for the pre-registered tests that we performed in the next section.

**Table 1. Descriptive statistics of Phase II participants**

|  | Overall (*N* = 491) |
|---|---|
| Gender |  |
|   Female | 252 (51.3%) |
|   Male | 239 (48.7%) |
| Age |  |
|   Mean (SD) | 46.8 (15.3) |
|   Median [Min, Max] | 48.0 [19.0, 81.0] |
| Technological affinity |  |
|   Mean (SD) | 4.26 (1.25) |
|   Median [Min, Max] | 4.33 [1.00, 7.00] |
| Regularity of LLM use |  |
|   Mean (SD) | 3.66 (2.03) |
|   Median [Min, Max] | 4.00 [1.00, 7.00] |
| Knowledge about LLMs |  |
|   Mean (SD) | 3.99 (1.73) |
|   Median [Min, Max] | 4.00 [1.00, 7.00] |
| Money burned in control block |  |
|   Mean (SD) | 0.0979 (0.223) |
|   Median [Min, Max] | 0 [0, 1.00] |
| Money burned in actual LLM use block, when use = 0 |  |
|   Mean (SD) | 0.0966 (0.229) |
|   Median [Min, Max] | 0 [0, 1.00] |
| Money burned in actual LLM use block, when use = 1 |  |
|   Mean (SD) | 0.137 (0.225) |
|   Median [Min, Max] | 0.0500 [0, 1.00] |
| Money burned in actual LLM use block, when use = 2 |  |
|   Mean (SD) | 0.181 (0.244) |
|   Median [Min, Max] | 0.100 [0, 1.00] |
| Money burned in actual LLM use block, when use = 3 |  |
|   Mean (SD) | 0.246 (0.284) |
|   Median [Min, Max] | 0.150 [0, 1.00] |
| Money burned in actual LLM use block, when use = 4 |  |
|   Mean (SD) | 0.296 (0.330) |
|   Median [Min, Max] | 0.200 [0, 1.00] |
| Money burned in actual LLM use block, when use = 5 |  |

|  | Overall (*N* = 491) |
|---|---|
| Mean (SD) | 0.358 (0.393) |
| Median [Min, Max] | 0.200 [0, 1.00] |
| Money burned in self-reported LLM use block, when use = 0 | |
| Mean (SD) | 0.115 (0.252) |
| Median [Min, Max] | 0 [0, 1.00] |
| Money burned in self-reported LLM use block, when use = 1 | |
| Mean (SD) | 0.149 (0.240) |
| Median [Min, Max] | 0.0500 [0, 1.00] |
| Money burned in self-reported LLM use block, when use = 2 | |
| Mean (SD) | 0.181 (0.250) |
| Median [Min, Max] | 0.100 [0, 1.00] |
| Money burned in self-reported LLM use block, when use = 3 | |
| Mean (SD) | 0.238 (0.287) |
| Median [Min, Max] | 0.100 [0, 1.00] |
| Money burned in self-reported LLM use block, when use = 4 | |
| Mean (SD) | 0.284 (0.331) |
| Median [Min, Max] | 0.150 [0, 1.00] |
| Money burned in self-reported LLM use block, when use = 5 | |
| Mean (SD) | 0.323 (0.377) |
| Median [Min, Max] | 0.150 [0, 1.00] |

### 4.2. Effect of access to an LLM

In **Table 2** we present estimates of a model that simultaneously compares actual and self-reported LLM use to the referential scenario in which someone doesn't have access to the (appropriate) LLM. The *OR* in all positive-use cases exceeds one (the *log-odds* are positive) and is statistically significant. Thus, we find empirical support for Hypothesis 1. We illustrate how LLM use relates to each use case in **Fig. 3A**.

**Table 2. Access to a capable LLM and money burning behavior**

|  | Money burning | | | |
|---|---|---|---|---|
|  | *OR* | *95% CI* | *z* | *p* |
| Intercept | 0.13*** | 0.10 – 0.15 | -21.89 | <0.001 |
| Actual LLM use = 0 | 0.96 | 0.84 – 1.11 | -0.52 | 1.000 |
| Actual LLM use = 1 | 1.90*** | 1.65 – 2.19 | 8.76 | <0.001 |
| Actual LLM use = 2 | 2.30*** | 2.00 – 2.66 | 11.39 | <0.001 |
| Actual LLM use = 3 | 2.80*** | 2.43 – 3.24 | 14.03 | <0.001 |
| Actual LLM use = 4 | 3.14*** | 2.72 – 3.63 | 15.59 | <0.001 |
| Actual LLM use = 5 | 4.83*** | 4.20 – 5.56 | 22.11 | <0.001 |
| Self-reported LLM use = 0 | 1.13 | 0.98 – 1.30 | 1.67 | 0.382 |
| Self-reported LLM use = 1 | 1.98*** | 1.72 – 2.29 | 9.40 | <0.001 |
| Self-reported LLM use = 2 | 2.32*** | 2.01 – 2.68 | 11.49 | <0.001 |
| Self-reported LLM use = 3 | 2.73*** | 2.37 – 3.15 | 13.75 | <0.001 |
| Self-reported LLM use = 4 | 3.03*** | 2.63 – 3.50 | 15.23 | <0.001 |
| Self-reported LLM use = 5 | 4.02*** | 3.50 – 4.62 | 19.52 | <0.001 |
| Male participant | 0.77 | 0.62 – 0.97 | -2.27 | 0.116 |
| Age | 1.01 | 0.90 – 1.13 | 0.21 | 1.000 |
| Technological affinity | 1.10 | 0.98 – 1.24 | 1.58 | 0.382 |
| Regularity of LLM use | 0.85 | 0.76 – 0.97 | -2.52 | 0.070 |
|  | Random Effects | | | |
| $\sigma^2$ | 0.89 | | | |
| $\tau_{00\ \text{Participant}}$ | 1.41 | | | |
| $\tau_{00\ \text{Block order}}$ | 0.00 | | | |
| ICC | 0.61 | | | |
| $N_{\text{Participant}}$ | 491 | | | |
| Observations (13 per participant) | 6383 | | | |
| Marginal $R^2$ / Conditional $R^2$ | 0.108 / 0.656 | | | |

*Notes: Age*, *Technological affinity* and *Regularity of LLM use* are centered to allow interpretability of the intercept: the intercept corresponds to money burning of female participants at mean levels of these variables.

* $p < 0.05$, ** $p < 0.01$, *** $p < 0.001$

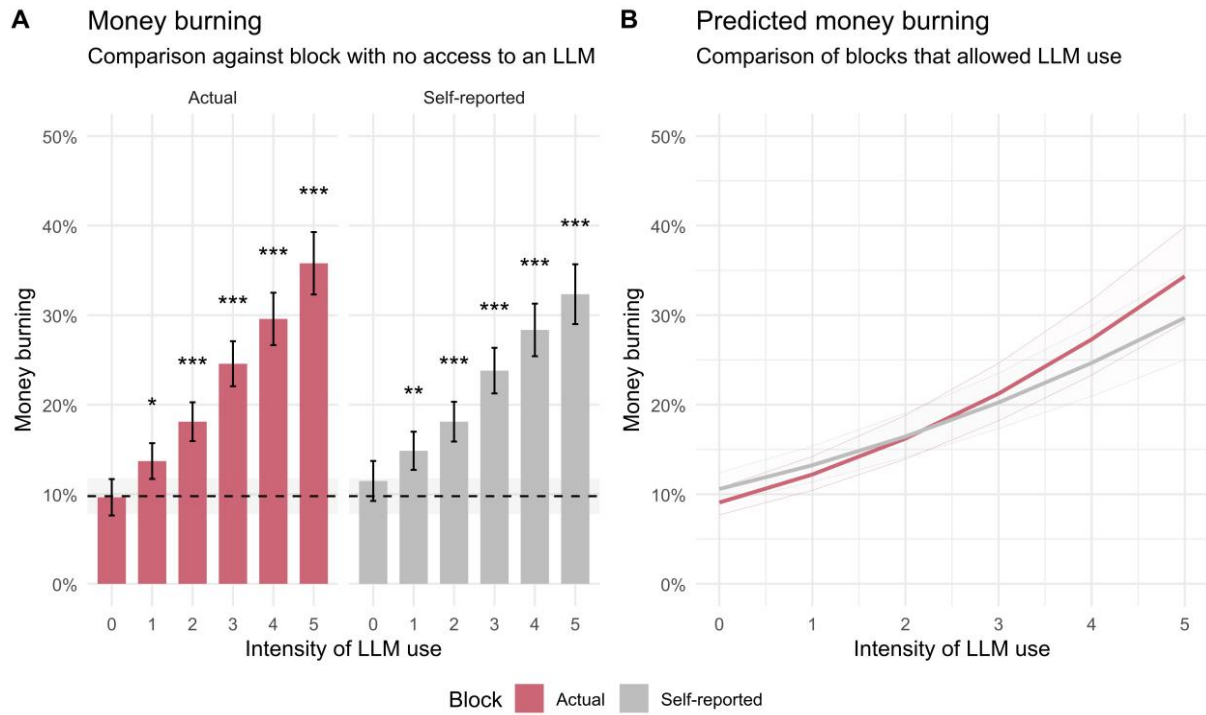

**Fig. 3. Antisocial behavior across various LLM use-cases (Panel A) and predicted antisocial behavior depending on intensity of actual and self-reported LLM use (Panel B)**

*Notes:* In **Panel A**, the dashed line corresponds to mean money burning against participants that did not have the opportunity to use the LLM (control scenario), and the shaded area corresponds to the 95% *CI*. ***, ** and * show statistically significant difference between the highlighted scenario and the control scenario (Welch's *t*-test with Holm-Bonferroni correction for *p*-values) at the 0.1, 1, and 5% level, respectively. In **Panel B** shaded areas correspond to 95% *CIs*.

### 4.3. Intensity of actual LLM-use, null use, and difference between and actual and self-reported use

The former analysis made a series of comparisons of various use cases against the scenario in which a person didn't have access to a capable LLM and thus had no possibility to "take shortcuts". We complement this with a regression run on participants in the *Actual LLM use* and *Self-reported LLM use* block, that simultaneously tests three hypotheses concerning the intensity of LLM use and antisocial behavior. Estimates are presented in **Table 3**.

**Table 3. Intensity of LLM use and money burning**

|  | Money burning | | | |
|---|---|---|---|---|
|  | *OR* | *95% CI* | *z* | *p* |
| Intercept | 0.10*** | 0.08 – 0.12 | -24.88 | <0.001 |
| Intensity | 1.39*** | 1.34 – 1.45 | 15.85 | <0.001 |
| Self-reported | 1.19*** | 1.08 – 1.30 | 3.61 | <0.001 |
| Intensity × Self-reported | 0.93*** | 0.90 – 0.95 | -4.90 | <0.001 |
| Male participant | 0.72** | 0.56 – 0.92 | -2.59 | 0.010 |
| Age | 1.08 | 0.95 – 1.23 | 1.22 | 0.223 |
| Technological affinity | 1.07 | 0.94 – 1.23 | 1.03 | 0.304 |
| Regularity of LLM use | 0.92 | 0.80 – 1.05 | -1.21 | 0.225 |
| Random Effects | | | | |
| $\sigma^2$ | 1.15 | | | |
| $\tau_{00}$ Participant | 1.67 | | | |
| $\tau_{00}$ Block order | 0.00 | | | |
| $\tau_{11}$ Participant/intensity | 0.15 | | | |
| $\rho_{01}$ Participant | -0.14 | | | |
| ICC | 0.70 | | | |
| N Participant | 491 | | | |
| Observations (12 per participant) | 5892 | | | |
| Marginal $R^2$ / Conditional $R^2$ | 0.071 / 0.724 | | | |

*Notes:* Self-reported is an indicator variable taking the value of 1 when use is self-reported, and 0 when it is observable. Age, Technological affinity and Regularity of LLM use are centered to allow interpretability of the intercept: the intercept corresponds to money burning of female participants at mean levels of these variables.

* $p < 0.05$, ** $p < 0.01$, *** $p < 0.001$

We now take the reader through various parts of our analysis. Our model was constructed, so that the coefficient for *Intensity* measured the change in antisocial behavior as *actual* LLM use increases (i.e., *Self-reported* = 0). This *OR* (*log-odds*) is larger than one (positive) and statistically significant, which indicates that antisocial behavior indeed increases in actual use cases. This supports Hypothesis 2.

The coefficient for *Intensity × Self-reported* measures how the intensity effect is different between self-reported and actual use. The *OR* (*log-odds*) is smaller than one (negative) and statistically significant, which suggests that increased self-reported use increases antisocial behavior to a smaller extent than increased actual use. Thus, Hypothesis 4 is supported.

Finally, the coefficient for *Self-reported* measures how antisocial behavior differs between the case when self-reported use is zero, and when actual use is zero. The larger than one *OR* (positive *log-odds*) and statistically significant coefficient indicates that antisocial behavior against people with null use is greater when they self-report null use, rather than when their actual use is null. This is consistent with Hypothesis 3.

For an illustration of the relationship between LLM use and antisocial behavior – separately for actual and self-reported use – see **Fig. 3B**.

### 4.4. Robustness analysis

We have also performed an analysis of the data relying on the pre-registered (strict) exclusion criterion: excluding participants that did not correctly answer *either* of the two attention checks. Results are shown in Appendix Tables A1 and A2. As indicated previously, results are qualitatively the same.

Moreover, given that it is possible that burning behavior in earlier scenarios affected behavior in later scenarios, as a robustness test, we have done a separate analysis on solely behavior in the first of the three blocks that were shown. The results are qualitatively the same (see Tables A1-2 in the Appendix).

### 4.5. Exploratory analyses

#### 4.5.1. Moderators

In **Fig. 4** and Table A3 in the Appendix we show how the exacerbation of punishment differs across people with differing levels of technological affinity, knowledge about LLMs and regularity of use of LLMs. The more regularly people use LLMs, the less their punishing behavior increased with the intensity of use. Punishing behavior did not increase more for people with higher technological affinity or self-reported knowledge about LLMs.

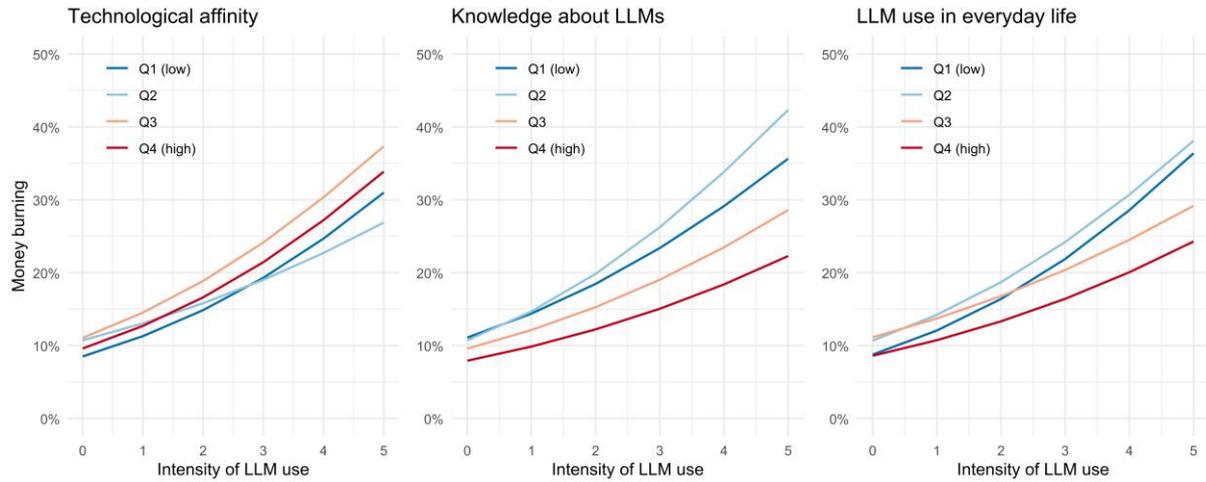

**Fig. 4. Moderators of antisocial behavior**

*Notes:* We used quartiles to enhance clarity. See Appendix Table A3 for more specific tests.

### 4.5.2. Correlations with measures of character

Finally, we answer to what extent (1) LLM use correlates with character judgments (Reif et al., 2025), and (2) character judgments correlate with money burning. We computed a *Laziness* aggregate score, which was based on four items, and a *Competence* score, also computed based on four items. However, we have to point out that by design, our focus in the experiment was to study behavior against LLM users and not impressions of them. The correlations discussed in this subsection could reflect the need for consistency between behavior and character judgments.

The higher the intensity of LLM use, the less competent the people that used them were rated, likewise for actual use ($r = -0.49$, 95% *CI* [-0.55, -0.42], $t(495) = -12.40$, $p < .001$) and self-reported use ($r = -0.40$, 95% *CI* [-0.48, -0.33], $t(495) = -9.84$, $p < .001$). Conversely, LLM use was positively correlated with laziness ratings, similarly for actual ($r = 0.49$, 95% *CI* [0.42, 0.55], $t(495) = 12.39$, $p < .001$) and self-reported use ($r = 0.39$, 95% *CI* [0.31, 0.46], $t(495) = 9.47$, $p < .001$).

Looking at people whose LLM use was observable, there was a negative correlation between competence and money burning behavior ($r = -0.37$, 95% CI [-0.44, -0.29], $t(495) = -8.79$, $p < .001$) and a positive correlation between laziness and money burning behavior ($r = 0.43$, 95% *CI* [0.35, 0.50], $t(495) = 10.50$, $p < .001$).

## 5. Discussion

Our study provides the first behavioral evidence that individuals are willing to incur personal costs to punish peers for using LLMs. Extending previous attitudinal research (Cheong et al., 2025; Reif et al., 2025) which documents that AI users are perceived as less competent and more lazy, we demonstrate that such negative evaluations translate into costly antisocial behavior. On average, participants destroyed 36% of the earnings of individuals who exclusively relied on LLMs to complete the task, indicating that disapproval of AI use is sufficiently strong to elicit punitive action rather than mere disapproval.

Consistent with theories of inequity aversion and effort-based deservingness (Fehr & Schmidt, 1999; Yang et al., 2016), punishment increased monotonically with the intensity of LLM use. Heavy reliance on LLM was treated as a more severe violation of the effort–reward principle (Feather, 1999; Konow, 2003), echoing findings that reputational penalties scale with the perceived extent of shortcut-taking (Acar et al., 2025.; Falk et al., 2005). Importantly, these results suggest that individuals are not only sensitive to whether LLM is used but also to how much it is used, reinforcing the idea that fairness norms operate in a graded rather than binary manner.

Second, and more novel, our findings illuminate a critical signaling dilemma in environments of imperfect information. Grounded in signaling theory (Connelly et al., 2011; Spence, 1973), we discovered that individuals who self-reported LLM use were punished more harshly than those whose LLM use was directly observable - as long as no use or low use was reported. However, as the intensity of use increased, the pattern reversed: punishment increased at a steeper rate for *actual* use than for self-reported use. As others have suggested (Jungbauer & Waldman, 2023; Gonçalves et al., 2025), a declaration of low-use and especially null-use may not be seen as a virtuous signal of effort, but rather as a non-credible, and potentially dishonest, claim aimed at avoiding stigma—a form of suspected under-reporting. At the same time, declarations of high-use may be more likely taken at face value, and do not suffer from an increased punishment due to a suspected misrepresentation. This credibility gap complicates the transparency dilemma (Draxler et al., 2024; Schilke & Reimann, 2025): although transparency is promoted as an ethical imperative, our findings show that disclosure can backfire, exposing individuals to social sanctions even when they tell the truth. Even "pure" human effort is subject to greater suspicion and harsher sanction once human-machine collaboration becomes possible, because observers cannot rule out the possibility of concealed LLM use, creating a state of ambient suspicion. This echoes broader research showing that the presence of a potential shortcut contaminates perceptions of effort and fairness (Mazar et al., 2008; Gino et al., 2009). Once LLM assistance becomes a salient option, even genuine human effort in environment of imperfect information is discounted, as null-use may be disbelieved by others. This exposes even honest individuals to social sanctions, undermining the credibility and perceived legitimacy of unaided work.

Interestingly, and perhaps counter-intuitively, we could not find any evidence that technological affinity or knowledge about LLM systems influenced Money Burning behavior. Education or affinity with technology does not seem to shape perceptions of its legitimate use in our case of human-machine collaboration. Instead, the violation of the effort-reward principle appears to trigger a punitive response that is broadly shared, even among those who share positive opinions of the technology themselves. On the other hand, the regular use of LLM systems appears to lower punitive behavior, leading to less harsh punishment even when LLM use increases. Additionally, our analyses did indicate that perceptions of *laziness* and *competence* are tightly coupled with punishment behavior. Higher levels of LLM use correlated with lower competence ratings and higher laziness ratings, which in turn predicted greater willingness to burn money. This echoes the results of Reif et al. (2025) and provides convergent evidence that antisocial behavior is not arbitrary but rooted in moralized judgments about effort, legitimacy, and deservingness (Earp et al., 2024; Lerner, 1980).

This might imply that antisocial behavior towards LLM users is primarily driven by perceptions of the *user* and not of the *machine*. Taken together, these results suggest that punishment is directed less at the technology itself than at the character inferences drawn about those who use it. While further research is needed, this points to the central role of social perceptions – rather than technical literacy – in shaping punitive responses to LLM use.

In conclusion, our findings reveal a double bind for human agency in the age of AI: the very tools offered to enhance cognitive efficiency trigger costly social punishment when used, while the mechanisms for transparency (self-reporting) can exacerbate rather than alleviate distrust. This tension places individuals in an untenable position, where neither unaided effort nor transparent LLM collaboration guarantees social legitimacy. They are socially punished for using efficient tools, yet their self-reported pure human effort is met with suspicion. The result is a lose-lose dynamic that threatens to erode trust and undermine the benefits of human-AI collaboration.

## 5.1. Limitations

Several limitations of this study should be noted. First, our experiment employed a stylized real-effort task – emoji counting – which allowed for precise manipulation of AI assistance but may not capture the complexity of tasks where LLMs are typically deployed, such as writing, reasoning, or problem-solving. This controlled design ensures internal validity and was deliberately chosen to be on the edge of LLM capabilities at the time of the study. However, this specificity limits the generalizability of our findings. The social penalties observed may differ for tasks that are more creative, strategic, or deeply personal, where the human contribution is more distinct and potentially more highly valued. Future research should investigate whether these punitive responses persist or transform in contexts where AI acts as a collaborator on open-ended problems rather than a solver of well-defined tasks, as much research has already shown that AI and AI use is evaluated differently in different domains (Bellaiche et al., 2023; Cunningham et al., 2025; Magni et al., 2024).

Second, although we recruited a broadly representative UK sample, participants were drawn from Prolific. Crowdsourcing samples are known to differ in important ways from the general population, including higher digital literacy and greater familiarity with online tasks. Moreover, as prior research suggests that participants sometimes display in-group bias toward fellow crowdworkers, our results may represent a conservative estimate of the willingness to punish LLM use (see Abbink and Herrmann, 2011).

Third, our operationalization of antisocial behavior through the Money Burning Game, while a validated economic paradigm, captures only one dimension of a potential social response. It measures a direct, costly, and punitive action. In real-world settings, disapproval of AI use may manifest in more subtle ways, such as reduced trust, decreased cooperation, lower performance evaluations, or exclusion from collaborative projects. These less overt but equally damaging forms of social sanction were not measured here and represent a critical avenue for future study.

Finally, we did not give the self-reported LLM use condition the option *not* to disclose. In real world settings, an individual trying to conceal their AI use may be more likely to simply not mention having used LLMs at all, instead of making open statements of not having used an AI. Consequently, the signal of null-use in our study might have been considered especially 'loud' and invited even more suspicion, considering that participants who claimed not to have used LLMs were randomly sandwiched between admitting LLM users. It is therefore not to be denied, that some form of spillover might have taken place, as the salience of LLM use was much higher than it might be in real world settings.

## 6. Conclusion

This study provides the first behavioral evidence that individuals are willing to incur personal costs to punish peers for using large language models, demonstrating that the integration of large language models into human work is not merely a technical or economic shift, but a profound social one. Extending prior attitudinal work, we show that disapproval of AI use translates into costly antisocial behavior that scales with the intensity of use. Importantly, the results reveal a credibility gap: individuals who self-report low use of LLM are punished more harshly than those whose low use is directly observable, highlighting the paradox that transparency can undermine rather than restore trust.

Taken together, these findings point to a double bind for human agency in the age of AI: reliance on AI triggers sanction, while unaided effort invites suspicion. The result is a lose–lose dynamic in which neither transparency nor restraint guarantees social legitimacy. This tension underscores the need to design disclosure systems, cultural narratives, and governance frameworks that align social incentives with technological realities. More broadly, our results caution that the success of human–AI collaboration will depend not only on technological performance, but equally on the evolution of social norms that determine when and how such collaboration is deemed acceptable.

# Appendix

**Antisocial behavior towards large language model users: experimental evidence**

**Supplementary Tables**

**Table A1. Regressions on alternative samples – Model 1**

|  | Final sample | | | | Sample using strict (pre-registered) exclusions | | | | Sample using data from first experimental block | | | |
| --- | --- | --- | --- | --- | --- | --- | --- | --- | --- | --- | --- | --- |
|  | OR | 95% CI | z | p | OR | 95% CI | z | p | OR | 95% CI | z | p |
| Intercept | 0.13 *** | 0.10 – 0.15 | -21.89 | <0.001 | 0.10 *** | 0.08 – 0.13 | -17.94 | <0.001 | 0.13 *** | 0.11 – 0.15 | -21.69 | <0.001 |
| Actual LLM use = 0 | 0.96 | 0.84 – 1.11 | -0.52 | 1.000 | 0.90 | 0.75 – 1.09 | -1.07 | 1.000 | 0.96 | 0.79 – 1.16 | -0.46 | 1.000 |
| Actual LLM use = 1 | 1.90 *** | 1.65 – 2.19 | 8.76 | <0.001 | 1.77 *** | 1.46 – 2.14 | 5.88 | <0.001 | 1.82 *** | 1.49 – 2.21 | 6.01 | <0.001 |
| Actual LLM use = 2 | 2.30 *** | 2.00 – 2.66 | 11.39 | <0.001 | 2.22 *** | 1.84 – 2.69 | 8.26 | <0.001 | 2.18 *** | 1.79 – 2.64 | 7.84 | <0.001 |
| Actual LLM use = 3 | 2.80 *** | 2.43 – 3.24 | 14.03 | <0.001 | 2.83 *** | 2.35 – 3.43 | 10.79 | <0.001 | 2.74 *** | 2.26 – 3.34 | 10.09 | <0.001 |
| Actual LLM use = 4 | 3.14 *** | 2.72 – 3.63 | 15.59 | <0.001 | 3.20 *** | 2.64 – 3.87 | 12.00 | <0.001 | 3.04 *** | 2.50 – 3.69 | 11.21 | <0.001 |
| Actual LLM use = 5 | 4.83 *** | 4.20 – 5.56 | 22.11 | <0.001 | 4.77 *** | 3.96 – 5.74 | 16.58 | <0.001 | 4.50 *** | 3.72 – 5.43 | 15.57 | <0.001 |
| Self-reported LLM use = 0 | 1.13 | 0.98 – 1.30 | 1.67 | 0.382 | 1.11 | 0.92 – 1.34 | 1.09 | 1.000 | 1.17 | 0.98 – 1.40 | 1.72 | 0.344 |

|  | Final sample | | | | Sample using strict (pre-registered) exclusions | | | | Sample using data from first experimental block | | | |
|---|---|---|---|---|---|---|---|---|---|---|---|---|
|  | *OR* | *95% CI* | *z* | *p* | *OR* | *95% CI* | *z* | *p* | *OR* | *95% CI* | *z* | *p* |
| Self-reported LLM use = 1 | 1.98 *** | 1.72 – 2.29 | 9.40 | <0.001 | 1.98 *** | 1.64 – 2.39 | 7.09 | <0.001 | 1.94 *** | 1.62 – 2.33 | 7.12 | <0.001 |
| Self-reported LLM use = 2 | 2.32 *** | 2.01 – 2.68 | 11.49 | <0.001 | 2.26 *** | 1.87 – 2.73 | 8.44 | <0.001 | 2.24 *** | 1.86 – 2.69 | 8.56 | <0.001 |
| Self-reported LLM use = 3 | 2.73 *** | 2.37 – 3.15 | 13.75 | <0.001 | 2.76 *** | 2.28 – 3.33 | 10.55 | <0.001 | 2.55 *** | 2.12 – 3.07 | 9.96 | <0.001 |
| Self-reported LLM use = 4 | 3.03 *** | 2.63 – 3.50 | 15.23 | <0.001 | 3.10 *** | 2.57 – 3.74 | 11.80 | <0.001 | 2.90 *** | 2.42 – 3.49 | 11.36 | <0.001 |
| Self-reported LLM use = 5 | 4.02 *** | 3.50 – 4.62 | 19.52 | <0.001 | 3.93 *** | 3.26 – 4.73 | 14.46 | <0.001 | 3.96 *** | 3.31 – 4.74 | 15.01 | <0.001 |
| Male participant | 0.77 | 0.62 – 0.97 | -2.27 | 0.116 | 0.80 | 0.59 – 1.09 | -1.43 | 0.763 | 0.80 | 0.64 – 0.99 | -2.01 | 0.231 |
| Age | 1.01 | 0.90 – 1.13 | 0.21 | 1.000 | 1.04 | 0.89 – 1.21 | 0.46 | 1.000 | 1.02 | 0.91 – 1.14 | 0.38 | 1.000 |
| Technological affinity | 1.10 | 0.98 – 1.24 | 1.58 | 0.382 | 1.01 | 0.85 – 1.21 | 0.16 | 1.000 | 1.09 | 0.97 – 1.23 | 1.45 | 0.440 |
| Regularity of LLM use | 0.85 | 0.76 – 0.97 | -2.52 | 0.070 | 0.82 | 0.69 – 0.97 | -2.35 | 0.114 | 0.88 | 0.78 – 0.99 | -2.07 | 0.231 |
| Random Effects | | | | | | | | | | | | |
| $\sigma^2$ | 0.89 | | | | 0.93 | | | | 0.89 | | | |
| $\tau_{00}$ | 1.41 $_{id}$ | | | | 1.41 $_{id}$ | | | | 1.27 $_{id}$ | | | |
|  | 0.00 $_{first\ block\ shown}$ | | | | 0.00 $_{first\ block\ shown}$ | | | | | | | |
| ICC | 0.61 | | | | 0.60 | | | | 0.59 | | | |
| N | 491 | | | | 267 | | | | 491 | | | |
| Observations | 6383 | | | | 3471 | | | | 3437 | | | |

|  | Final sample | | | | Sample using strict (pre-registered) exclusions | | | | Sample using data from first experimental block | | | |
|---|---|---|---|---|---|---|---|---|---|---|---|---|
|  | OR | 95% CI | z | p | OR | 95% CI | z | p | OR | 95% CI | z | p |
| Marginal R² / Conditional R² | 0.108 / 0.656 | | | | 0.116 / 0.648 | | | | 0.111 / 0.634 | | | |

Notes: Age, Technological affinity and Regularity of LLM use are centered. * p<0.05** p<0.01*** p<0.001

**Table A2. Regressions on alternative samples – Model 2**

|  | Final sample | | | | Sample using strict (pre-registered) exclusions | | | | Sample using data from first experimental block | | | |
|---|---|---|---|---|---|---|---|---|---|---|---|---|
| *Predictors* | OR | 95% CI | z | p | OR | 95% CI | z | p | OR | 95% CI | z | p |
| Intercept | 0.10 *** | 0.08 – 0.12 | -24.88 | <0.001 | 0.06 *** | 0.05 – 0.08 | -22.28 | <0.001 | 0.07 *** | 0.05 – 0.09 | -19.58 | <0.001 |
| Intensity | 1.39 *** | 1.34 – 1.45 | 15.85 | <0.001 | 1.42 *** | 1.35 – 1.50 | 12.71 | <0.001 | 1.47 *** | 1.38 – 1.58 | 11.18 | <0.001 |
| Self-reported | 1.19 *** | 1.08 – 1.30 | 3.61 | <0.001 | 1.30 *** | 1.15 – 1.46 | 4.22 | <0.001 | 1.26 | 0.93 – 1.70 | 1.49 | 0.135 |
| Intensity × Self-reported | 0.93 *** | 0.90 – 0.95 | -4.90 | <0.001 | 0.90 *** | 0.86 – 0.94 | -5.22 | <0.001 | 0.90 * | 0.82 – 0.99 | -2.13 | 0.033 |
| Male participant | 0.72 ** | 0.56 – 0.92 | -2.59 | 0.010 | 0.76 | 0.55 – 1.06 | -1.63 | 0.104 | 0.68 * | 0.50 – 0.91 | -2.56 | 0.010 |
| Age | 1.08 | 0.95 – 1.23 | 1.22 | 0.223 | 1.13 | 0.95 – 1.34 | 1.40 | 0.161 | 1.10 | 0.95 – 1.28 | 1.28 | 0.200 |
| Technological affinity | 1.07 | 0.94 – 1.23 | 1.03 | 0.304 | 0.99 | 0.82 – 1.20 | -0.08 | 0.936 | 1.11 | 0.94 – 1.30 | 1.22 | 0.222 |
| Regularity of LLM use | 0.92 | 0.80 – 1.05 | -1.21 | 0.225 | 0.96 | 0.80 – 1.16 | -0.42 | 0.675 | 0.88 | 0.74 – 1.03 | -1.56 | 0.119 |

| Predictors | Final sample | | | | Sample using strict (pre-registered) exclusions | | | | Sample using data from first experimental block | | | |
|---|---|---|---|---|---|---|---|---|---|---|---|---|
| | OR | 95% CI | z | p | OR | 95% CI | z | p | OR | 95% CI | z | p |
| Random Effects | | | | | | | | | | | | |
| $\sigma^2$ | 1.15 | | | | 1.29 | | | | 0.99 | | | |
| $\tau_{00}$ | 1.67 id | | | | 1.52 id | | | | 2.41 id | | | |
| | 0.00 first block shown | | | | 0.00 first block shown | | | | | | | |
| $\tau_{11}$ | 0.15 id.intensity | | | | 0.15 id.intensity | | | | 0.22 id.intensity | | | |
| $\rho_{01}$ | -0.14 id | | | | -0.01 id | | | | -0.17 id | | | |
| ICC | 0.70 | | | | 0.69 | | | | 0.79 | | | |
| N | 491 | | | | 267 | | | | 491 | | | |
| Observations | 5892 | | | | 3204 | | | | 2946 | | | |
| Marginal $R^2$ / Conditional $R^2$ | 0.071 / 0.724 | | | | 0.070 / 0.709 | | | | 0.077 / 0.810 | | | |

Notes: Age, Technological affinity and Regularity of LLM use are centered. * p<0.05** p<0.01*** p<0.001

**Table A3. Technological affinity, LLM use, and LLM knowledge as moderators – Model 2**

| | Interaction between Intensity and Technological affinity | | | | Interaction between Intensity and Knowledge about LLMs | | | | Interaction between Intensity and Regularity of LLM use | | | |
|---|---|---|---|---|---|---|---|---|---|---|---|---|
| | OR | 95% CI | z | p | OR | 95% CI | z | p | OR | 95% CI | z | p |
| (Intercept) | 0.11 *** | 0.09 – 0.13 | -24.84 | <0.001 | 0.11 *** | 0.09 – 0.13 | -25.29 | <0.001 | 0.11 *** | 0.09 – 0.13 | -25.26 | <0.001 |
| Intensity | 1.34 *** | 1.29 – 1.39 | 15.16 | <0.001 | 1.34 *** | 1.29 – 1.39 | 15.20 | <0.001 | 1.34 *** | 1.29 – 1.39 | 15.28 | <0.001 |
| Technological affinity | 1.04 | 0.91 – 1.18 | 0.54 | 0.588 | | | | | | | | |
| Intensity × Technological affinity | 1.01 | 0.97 – 1.04 | 0.30 | 0.767 | | | | | | | | |
| Knowledge about LLMs | | | | | 0.92 | 0.80 – 1.05 | -1.29 | 0.197 | | | | |
| Intensity × Knowledge about LLMs | | | | | 0.97 | 0.94 – 1.01 | -1.43 | 0.152 | | | | |
| Regularity of LLM use | | | | | | | | | 0.99 | 0.87 – 1.12 | -0.17 | 0.862 |
| Intensity × Regularity of LLM use | | | | | | | | | 0.95 ** | 0.91 – 0.98 | -2.93 | 0.003 |
| Male participant | 0.72 ** | 0.56 – 0.92 | -2.64 | 0.008 | 0.76 * | 0.60 – 0.98 | -2.16 | 0.031 | 0.74 * | 0.58 – 0.94 | -2.43 | 0.015 |
| Age | 1.11 | 0.98 – 1.25 | 1.66 | 0.096 | 1.07 | 0.94 – 1.21 | 1.02 | 0.306 | 1.09 | 0.96 – 1.23 | 1.35 | 0.178 |
| Random Effects | | | | | | | | | | | | |
| $\sigma^2$ | 1.15 | | | | 1.15 | | | | 1.15 | | | |
| $\tau_{00}$ | 1.66 id | | | | 1.65 id | | | | 1.66 id | | | |
| | | | | | 0.00 first block shown | | | | 0.00 first block shown | | | |
| $\tau_{11}$ | 0.15 id.intensity | | | | 0.15 id.intensity | | | | 0.15 id.intensity | | | |
| $\rho_{01}$ | -0.13 id | | | | -0.14 id | | | | -0.13 id | | | |
| ICC | 0.70 | | | | 0.70 | | | | 0.70 | | | |
| N | 491 | | | | 491 | | | | 491 | | | |

|  | Interaction between Intensity and Technological affinity | | | | Interaction between Intensity and Knowledge about LLMs | | | | Interaction between Intensity and Regularity of LLM use | | | |
| --- | --- | --- | --- | --- | --- | --- | --- | --- | --- | --- | --- | --- |
|  | OR | 95% CI | z | p | OR | 95% CI | z | p | OR | 95% CI | z | p |
| Observations (12 per participant) | 5892 | | | | 5892 | | | | 5892 | | | |
| Marginal $R^2$ / Conditional $R^2$ | 0.069 / 0.723 | | | | 0.075 / 0.723 | | | | 0.078 / 0.723 | | | |

Notes: Age, Technological affinity, Knowledge about LLMs, and Regularity of LLM use are centered. * p<0.05** p<0.01*** p<0.001

# Experimental instructions

Below are the experimental instructions for the key part of the experiment – Phase II. A pdf printout of the instructions for both phases as available at the Open Science Framework at: https://osf.io/2yt9e/overview?view_only=8268c7ac95d94f2390fd3ed6dc76269a

## Phase II

**[Page 1]**

Hello! We're researchers from the [redacted] and the [redacted]. You're invited to take part in a study on how people perform when their work is evaluated by another person.

- **Duration**: Approximately 15 minutes.
- **Privacy**: This is an academic, not-for-profit project. Your responses will be fully anonymized and never linked to your identity in any reports or publications.

    **Questions?** Contact us at
    - [redacted]

You will complete a series of reviewing tasks, in which you will be evaluating the performance of other Prolific participants.

**By proceeding, you confirm that you:**
- Have been informed about how the study is conducted and why your participation is needed.
- Understand you may withdraw at any time without penalty.
- Know how your data will be processed under the [redacted] privacy policy, and understand your rights regarding your personal data.

Please click 'next' to start the study.

Please enter your Prolific ID below
[box]

**Survey Overview**

- This survey is different from what you may be used from other surveys on Prolific; What you do will be directly affecting **other users** of this platform. When this survey describes other people, the information you see is **real**, and has been collected at an earlier point in time on Prolific.

- The other participant has performed a series of counting tasks and has earned a potential bonus for doing so correctly. Your task will be to review their performance carefully.

- Below, you see an example of how the counting task looked like to the participants.

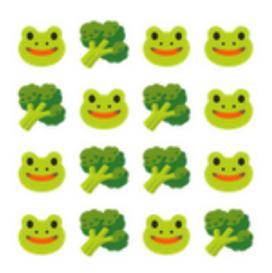

Over the series of 5 rounds, participants had to count different emojis as accurately as possible, with the tasks slowly increasing in difficulty.

Over the next pages, you will see each of the different counting tasks the other participants completed. Please carefully review them to have a feeling for the difficulty of the tasks. You do not need to count the requested emojis yourself.

Participants were asked to count the **mice** in this emoji grid.

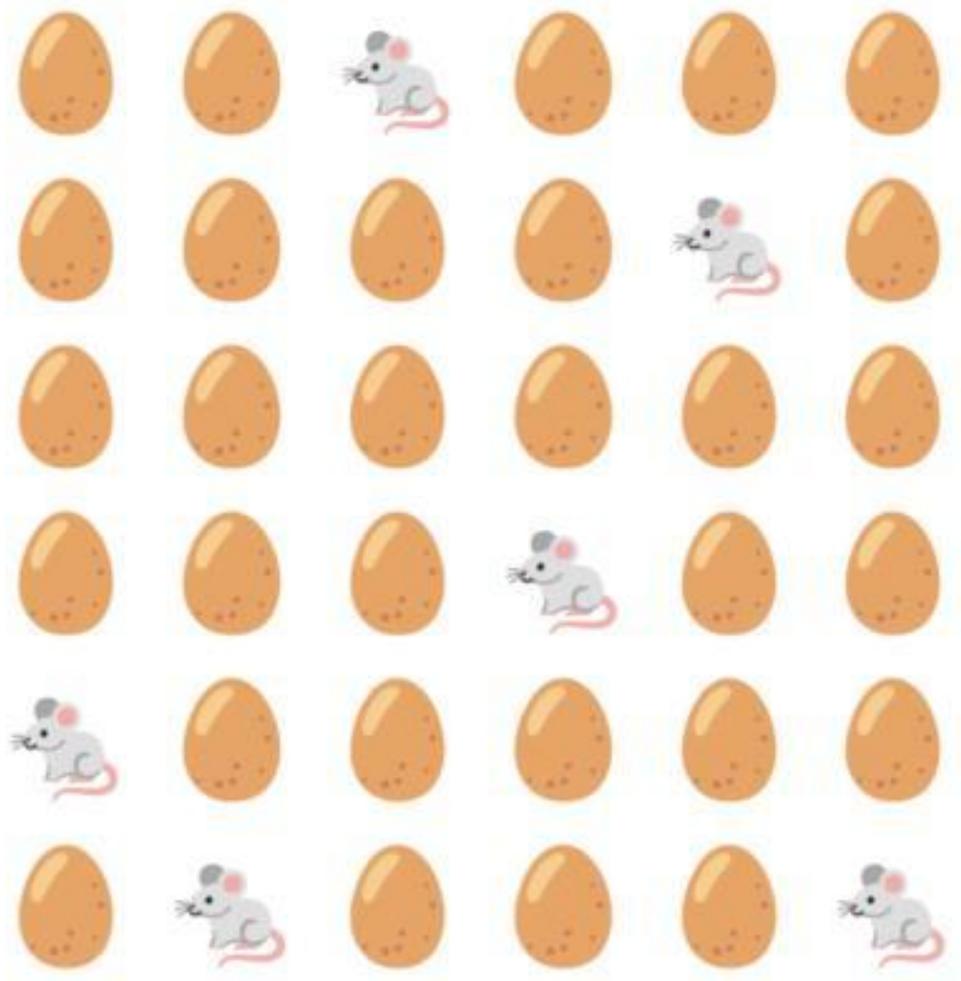

Participants were asked to count the **balloons** this emoji grid.

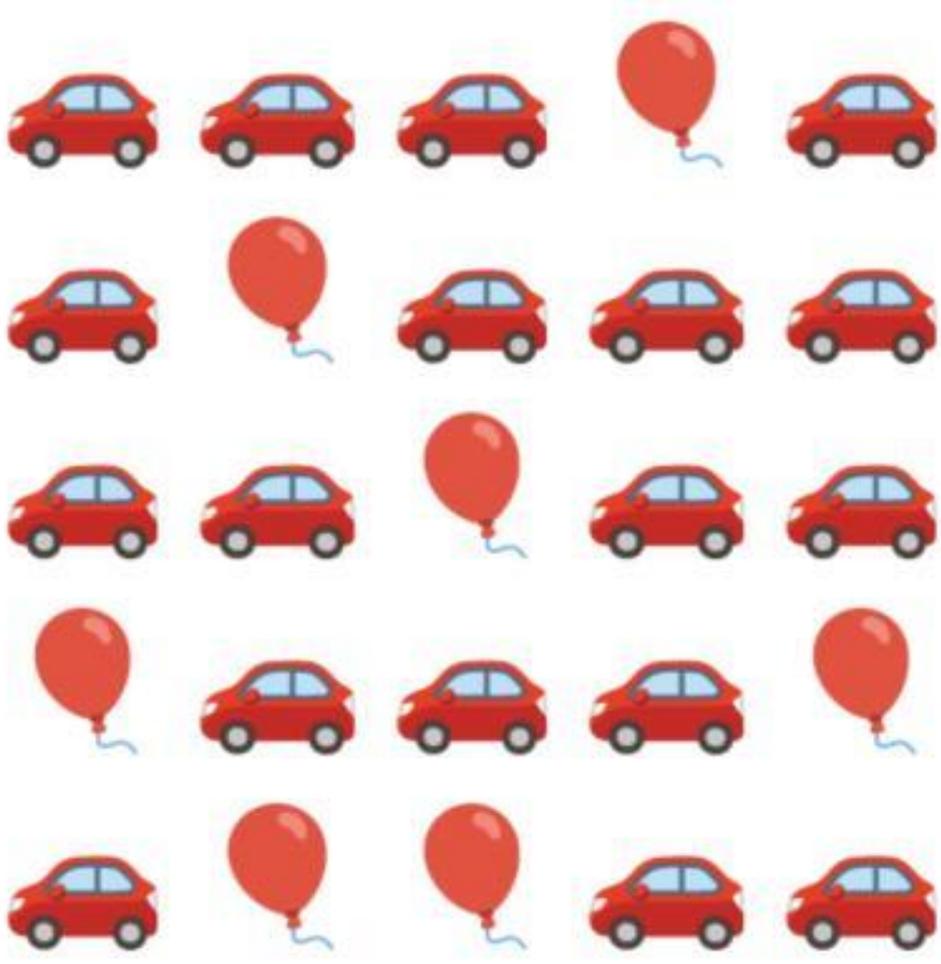

Participants were asked to count the **keys** in this emoji grid.

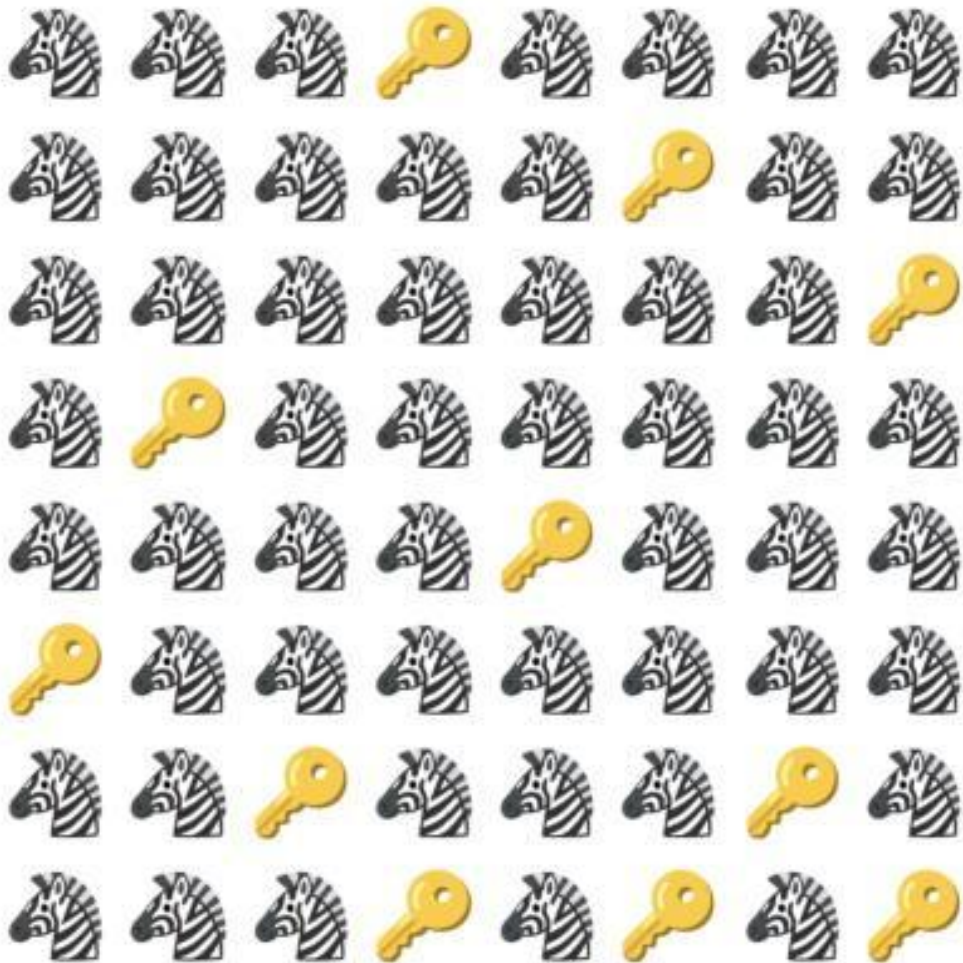

Participants were asked to count the **stars** in this emoji grid.

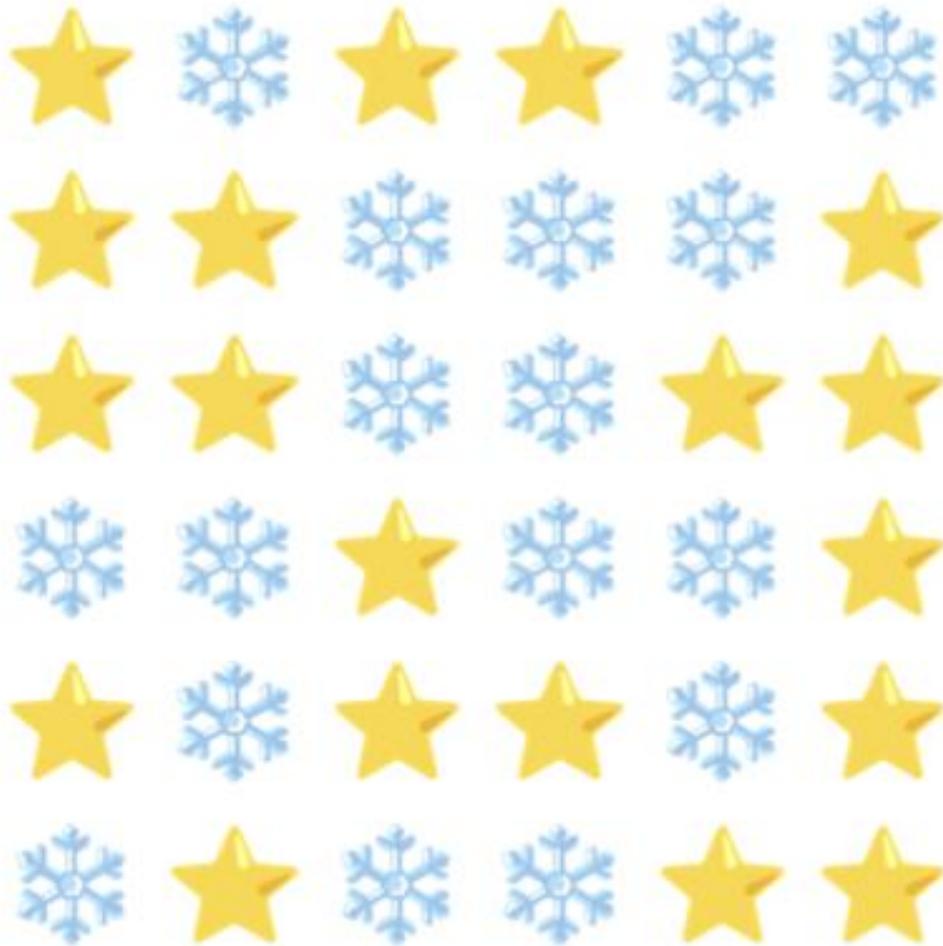

This is the last counting task that had to be completed.
Please click **Next**, once you have familiarized yourself with the task the other users had to solve. On the next page, your task will be explained in detail.

**Introduction to the Task**
- You will be given some information about the other participant's performance in the emoji counting task. Your task is to **carefully read** the information, and afterwards make a decision about the **real bonus** of the other participant.
- The other participant has received a potential endowment of 100 tokens (where 1 token = £*0.01*).
- ***You are now given an endowment of 40 tokens.*** This endowment is for you to keep and use. Unused tokens will be paid out to you after the end of the survey as a bonus payment.
- In this part of the study, you have the option to reduce the other participant's tokens by using some of your own tokens.

**How It Works**
- For every 1 token you spend, the other participant loses 5 tokens.
- *You do not gain anything from reducing the other participant's tokens.*
- You can choose any whole number of tokens between 0 and 20 to spend, allowing the other player to keep between 100 and 0 tokens.
- You will be asked to judge **13** different players. For each decision, your endowment resets back to 40. After the study is completed, a **random** decision out of these 13 will be chosen and performed -- for each decision you make, both your own endowment and the endowment of another participant can be affected. **Therefore, please make each decision as if it was the one that counts, as it very well might be!**

**Important Notes**
- The other participant cannot see your decision until the task is over.
- Your decision is completely anonymous — the other participant will not know who made it.
- There are no "right" or "wrong" choices. Please decide as you prefer.

*Below, you find an example of what the decision mechanisms looks like to get familiar with it. Once you choose a number in the list, the little calculator will live update to show the effect of this decision.*

**Example User**

> - Task: Emoji Counting (5 tasks)
> - Correct answers: 5/5
> - Earned endowment: 100 tokens

Please decide how many of your tokens you would like to spend to reduce the other participant's tokens.

Choose how many tokens you spend (0–20)

**Your choice:**

You spend **0** tokens — leaving you with **40** remaining tokens

The other player receives **100** tokens

Please move the slider to continue. If you want to select 0, please move the slider, and then move it back to 0.

Once you have read the instructions carefully and familiarized yourself with the decision menu, please click next to continue.

If you spend 2 of your tokens, by how many tokens is the other participant's endowment reduced?
- 2
- 5
- 10
- 20

How many of your decisions will count for real payments at the end?
- None
- 1 chosen at random
- 5 chosen at random
- All

Does your decision have real consequences?
- No – all decisions are purely hypothetical
- Yes – my own endowment gets impacted
- Yes – another player's endowment gets impacted
- Yes – both our endowment gets impacted

After each of the 13 decisions, how many tokens will you start with again?
- 40
- 100
- It depends on your previous decision
- Tokens always carry over

You will now be asked to make your first review. Please read about the performance of a random real user below:
**Random User**

- Task: Emoji Counting (5 tasks)
- Correct answers: 5/5
- Earned endowment: 100 tokens

Please decide how many tokens the other participant should be paid out.
Choose how many tokens you spend (0–20)
**Your choice:**
You spend **0** tokens — leaving you with **40** remaining tokens
The other player receives **100** tokens

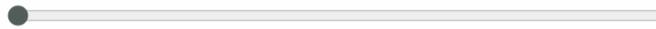

Please move the slider to continue. If you want to select 0, please move the slider, and then move it back to 0.
Please rate the extent to which you believe the following adjectives or phrases accurately describe this person, on a scale of 1 (does not describe this person at all) to 7 (does describe this person very well).

|  | Not at all 1 | 2 | 3 | 4 | 5 | 6 | Very well 7 |
|---|---|---|---|---|---|---|---|
| Capable | ○ | ○ | ○ | ○ | ○ | ○ | ○ |
| Skillful | ○ | ○ | ○ | ○ | ○ | ○ | ○ |
| Masterful | ○ | ○ | ○ | ○ | ○ | ○ | ○ |
| Gives Up Easily | ○ | ○ | ○ | ○ | ○ | ○ | ○ |
| Irresponsible | ○ | ○ | ○ | ○ | ○ | ○ | ○ |
| Lazy | ○ | ○ | ○ | ○ | ○ | ○ | ○ |
| Complacent | ○ | ○ | ○ | ○ | ○ | ○ | ○ |
| Competent | ○ | ○ | ○ | ○ | ○ | ○ | ○ |

Please read this information carefully now, as the next participants solved the same tasks with **some big differences**.

Unlike the task you saw before, these users here were given the option to receive support from a Large Language Model (LLM; for example, ChatGPT or Deepseek are LLMs), in solving the problem. Users were informed that the LLM-Support was generated through a commonly used Large Language Model and reliable. There was no price to pay or reward to receive from using the support -- users were completely free in deciding whether they wanted to use it or not.

The LLM input looked something like this:

> Would you like to view the LLM generated support for this task?
>
> ❓ Select all that apply
>    Please select at most one answer
>
> ☑ Yes

> This is the answer our chosen LLM language model has provided:
>
> I can count 9 frogs in the image.

The LLM-generated advice text only showed up after actively selecting it. By default, no support was given, and the users had to actively choose to turn it on for each individual question.

All following participants will have had access to this LLM-Support. However, two different groups exist - a **red group** and a **blue group**. The differences between those groups will be explained to you on the next couple of pages. You will be asked to review **both groups.** After reviewing them, we are going to ask you a few short demographic questions, before you will complete the survey.

Please answer these few short questions before going on to the review task.
There was a high chance that the LLM would give incorrect solutions to the users.
- True
- False

The so called 'LLM Advice' was actually hand-written by a human.
- True
- False

Please rate the extent to which you believe the following adjectives or phrases accurately describe this person, on a scale of 1 (does not describe this person at all) to 7 (does describe this person very well).

|  | Not at all | | | | | | Very well |
|---|---|---|---|---|---|---|---|
|  | 1 | 2 | 3 | 4 | 5 | 6 | 7 |
| Lazy | ○ | ○ | ○ | ○ | ○ | ○ | ○ |
| Irresponsible | ○ | ○ | ○ | ○ | ○ | ○ | ○ |
| Capable | ○ | ○ | ○ | ○ | ○ | ○ | ○ |
| Complacent | ○ | ○ | ○ | ○ | ○ | ○ | ○ |
| Gives up easily | ○ | ○ | ○ | ○ | ○ | ○ | ○ |
| Competent | ○ | ○ | ○ | ○ | ○ | ○ | ○ |
| Skillfull | ○ | ○ | ○ | ○ | ○ | ○ | ○ |
| Masterful | ○ | ○ | ○ | ○ | ○ | ○ | ○ |

You will now be explained the **red group**'s survey task.

As a reminder, this is what the LLM input looked like:

> Would you like to view the LLM generated support for this task?
>
> ❓ Select all that apply
>     Please select at most one answer
>
> ☑ Yes

> This is the answer our chosen LLM language model has provided:
>
> I can count 9 frogs in the image.

For the **red group**, we were able to collect **objective data** on whether they used the LLM to solve the task, by tracking whether they turned on the LLM-Support on each individual task. In the next reviews, you will also see **how often the participants used LLM to achieve their final score.** Please consider that information carefully when making your decisions.

You will now be shown the performance of six random users, who all had the ability to use the LLM-Support. Please carefully review the information you receive, and make a thoughtful choice for your review task.

For one randomly selected user, we will also ask a few questions about your perceptions of them.

### Random User

- Task: Emoji Counting (5 tasks)
- Correct answers: 5/5
- Earned endowment: 100 tokens
- **LLM Used: out of 5 times**

### Choose how many tokens you spend (0–20)

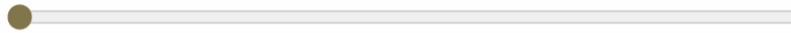

### Your choice:

You spend **0** tokens — leaving you with **40** remaining tokens

The other player receives **100** tokens

Please rate the extent to which you believe the following adjectives or phrases accurately describe this person, on a scale of 1 (does not describe this person at all) to 7 (does describe this person very well).

|  | Not at all | | | | | | Very well |
|---|---|---|---|---|---|---|---|
|  | 1 | 2 | 3 | 4 | 5 | 6 | 7 |
| Gives Up Easily | ○ | ○ | ○ | ○ | ○ | ○ | ○ |
| Complacent | ○ | ○ | ○ | ○ | ○ | ○ | ○ |
| Skillful | ○ | ○ | ○ | ○ | ○ | ○ | ○ |
| Masterful | ○ | ○ | ○ | ○ | ○ | ○ | ○ |
| Capable | ○ | ○ | ○ | ○ | ○ | ○ | ○ |
| Lazy | ○ | ○ | ○ | ○ | ○ | ○ | ○ |
| Irresponsible | ○ | ○ | ○ | ○ | ○ | ○ | ○ |
| Competent | ○ | ○ | ○ | ○ | ○ | ○ | ○ |

Please answer these few questions before carrying on with the survey.

The information about LLM use was collected by asking the users how often they used the LLM support.
- True
- False

How many tasks did the users get correct in the batch of performances you just reviewed?

- No information given
- Always different
- 3/5 correctly answered tasks
- 5/5 correctly answered tasks

The data on how often the LLM was used was objective.
- True
- False

**[Page 20] – Start of self-reported use ("Blue") Group**

You will now be explained the **blue group**'s survey task.

As a reminder, this is what the LLM input looked like:
The LLM input looked something like this:

[Screenshot: "Would you like to view the LLM generated support for this task? Select all that apply. Please select at most one answer. ☑ Yes"]

[Screenshot: "This is the answer our chosen LLM language model has provided: I can count 9 frogs in the image."]

The LLM generated advice text only showed up after actively selecting it. By default, no support was given, and the users had to actively choose to turn it on for each individual question.

These users, however, had to **self-disclaim** whether they used LLM or not. You will only see **how much the players were willing to share**, based on their own free choice and preference. No verification of this **self-reported data** took place. For the next six reviews, you will only see the self-disclaimer about these users' LLM use. Please consider that information carefully when making your decisions.

Here you can see the options of self-disclaiming the participants had.

[Screenshot showing options:
*How often have you used the LLM support to solve the counting tasks in this study?
Choose one of the following answers
○ I have not used the LLM support at all. (0 times)
○ I have used the LLM support 1 time.
○ I have used the LLM support 2 times.
○ I have used the LLM support 3 times.
○ I have used the LLM support 4 times.
○ I have used the LLM support all 5 times.]

**[Pages 21-26; six pages for each of the six use cases, presented in random order]**

You will now be shown the performance of six random users, who all had the ability to use the LLM-Support. Please carefully review the information you receive, and make a thoughtful choice for your review task.

For one randomly selected user, we will also ask a few questions about your perceptions of them.

**Random User**
- Task: Emoji Counting (5 tasks)
- Correct answers: 5/5
- Earned endowment: 100 tokens
- **The User reported: "I used the LLM 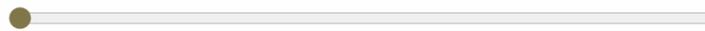 out of 5 times."**

**Choose how many tokens you spend (0–20)**

**Your choice:**
You spend **0** tokens — leaving you with **40** remaining tokens
The other player receives **100** tokens

Please rate the extent to which you believe the following adjectives or phrases accurately describe this person, on a scale of 1 (does not describe this person at all) to 7 (does describe this person very well).

Please rate the extent to which you believe the following adjectives or phrases accurately describe this person, on a scale of 1 (does not describe this person at all) to 7 (does describe this person very well).

|  | Not at all | | | | | | Very well |
|---|---|---|---|---|---|---|---|
|  | 1 | 2 | 3 | 4 | 5 | 6 | 7 |
| Complacent | ○ | ○ | ○ | ○ | ○ | ○ | ○ |
| Lazy | ○ | ○ | ○ | ○ | ○ | ○ | ○ |
| Skillfull | ○ | ○ | ○ | ○ | ○ | ○ | ○ |
| Competent | ○ | ○ | ○ | ○ | ○ | ○ | ○ |
| Masterful | ○ | ○ | ○ | ○ | ○ | ○ | ○ |
| Gives up easily | ○ | ○ | ○ | ○ | ○ | ○ | ○ |
| Irresponsible | ○ | ○ | ○ | ○ | ○ | ○ | ○ |
| Capable | ○ | ○ | ○ | ○ | ○ | ○ | ○ |

Which group did you just review?
- Blue group (self-disclaimed)
- Red group (objective data)

The information about LLM use was collected by asking the users how often they used the LLM support.
- True
- False

The data on how often the LLM was used was objective.
- True
- False

Thank you! You have completed all of the review tasks. We would kindly ask you to answer the following demographics questions.
To what extent do you disagree or agree with the following statement below?

Please rate on a scale of 1 (fully disagree) to 7 (fully agree).
To what extent do you disagree or agree with the following statement below?

| | Fully Disagree 1 | 2 | 3 | 4 | 5 | 6 | Fully Agree 7 |
|---|---|---|---|---|---|---|---|
| I like to occupy myself in greater detail with technical systems. | ○ | ○ | ○ | ○ | ○ | ○ | ○ |
| I like testing the functions of new technical systems. | ○ | ○ | ○ | ○ | ○ | ○ | ○ |
| I predominantly deal with technical systems because I have to. | ○ | ○ | ○ | ○ | ○ | ○ | ○ |
| When I have a new technical system in front of me, I try it out intensively. | ○ | ○ | ○ | ○ | ○ | ○ | ○ |
| I enjoy spending time becoming acquainted with a new technical system. | ○ | ○ | ○ | ○ | ○ | ○ | ○ |
| It is enough for me that a technical system works; I don't care how or why. | ○ | ○ | ○ | ○ | ○ | ○ | ○ |
| I try to understand how a technical system exactly works. | ○ | ○ | ○ | ○ | ○ | ○ | ○ |
| It is enough for me to know the basic functions of a technical system. | ○ | ○ | ○ | ○ | ○ | ○ | ○ |
| I try to make full use of the capabilities of a technical system. | ○ | ○ | ○ | ○ | ○ | ○ | ○ |

Please rate on a scale of 1 (fully disagree) to 7 (fully agree).

|  | Fully Disagree | | | | | | Fully Agree |
|---|---|---|---|---|---|---|---|
|  | 1 | 2 | 3 | 4 | 5 | 6 | 7 |
| I frequently use LLM systems. | ○ | ○ | ○ | ○ | ○ | ○ | ○ |
| I have a good understanding of the capacity of LLM systems. | ○ | ○ | ○ | ○ | ○ | ○ | ○ |
| I have a good understanding of limitations of LLM systems. | ○ | ○ | ○ | ○ | ○ | ○ | ○ |

If you have any comments for us researchers, please let us know here!

[Box]

**This is the Prolific Completion Code:**
If you have any further questions regarding this study, please feel free to contact either of the researchers conducting this study:
[redacted]

Once more, thank you for your participation and your contribution to the advancement of knowledge on human-machine interaction!